\pdfoutput=1

\documentclass[11pt]{article}
\usepackage[table]{xcolor}

\usepackage[]{acl}
\usepackage{amssymb}
\usepackage{amsmath}
\usepackage{times}
\usepackage{latexsym}
\usepackage{algorithm}
\usepackage{algpseudocode}
\usepackage{fancyvrb}
\usepackage{alltt,xcolor}
\usepackage{enumitem}
\definecolor{lightgray}{gray}{0.95}

\usepackage[most]{tcolorbox} 
\usepackage{xcolor} 

\newtcolorbox{FVerbatim}{
  colback=lightgray, 
  colframe=black, 
  boxsep=0pt, 
  top=10pt, 
  bottom=10pt, 
  left=10pt, 
  right=10pt, 
  arc=0pt, 
  boxrule=1pt, 
  breakable, 
  before upper=\begingroup\alltt, 
  after upper=\endgroup 
}

\usepackage[T1]{fontenc}

\usepackage[utf8]{inputenc}

\usepackage{microtype}
\usepackage{booktabs}
\usepackage{multirow}
\usepackage{tikz}
\usepackage{graphicx}
\usepackage{fancyvrb}

\title{Distillation Contrastive Decoding: Improving LLMs Reasoning with Contrastive Decoding and Distillation
}

\author{Phuc Phan$^\ast$, Hieu Tran\thanks{* Equal contribution} \and Long Phan \\
        VietAI Research}

\begin{document}
\maketitle
\begin{abstract}
We propose a straightforward approach called Distillation Contrastive Decoding (DCD) to enhance the reasoning capabilities of Large Language Models (LLMs) during inference. In contrast to previous approaches that relied on smaller \textit{amateur} models or analysis of hidden state differences, DCD employs Contrastive Chain-of-thought Prompting and advanced distillation techniques, including Dropout and Quantization. This approach effectively addresses the limitations of Contrastive Decoding (CD), which typically requires both an \textit{expert} and an \textit{amateur} model, thus increasing computational resource demands. By integrating contrastive prompts with distillation, DCD obviates the need for an amateur model and reduces memory usage. Our evaluations demonstrate that DCD significantly enhances LLM performance across a range of reasoning benchmarks, surpassing both CD and existing methods in the GSM8K and StrategyQA datasets.\footnote{Code is available at \url{https://github.com/pphuc25/distil-cd}}

\end{abstract}

\begin{figure*}[h]
\centering
\includegraphics[width=16cm]{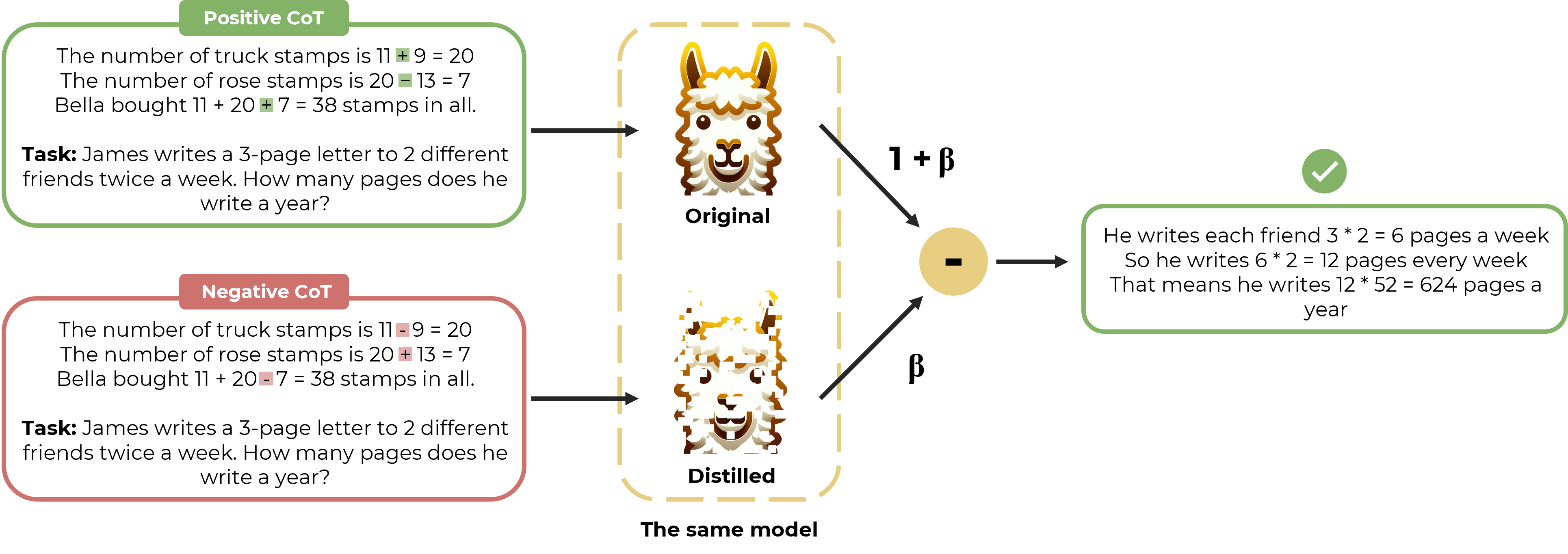}
\caption{An overview of Distillation Contrastive Decoding method. Valid CoT demonstrations as well as the query will be sent to an LLM, while invalid CoT demonstrations and the query will be sent into a distilled version of the model. We will then use this logit information to enhance the reasoning decoding process.}
\label{fig:figure1}
\end{figure*}

\section{Introduction}
Reasoning capabilities in LLMs refer to the models' ability to analyze, understand, and infer information, mirroring human-like logical reasoning. Recently, the reasoning skills of LLMs have seen substantial advancements, showcasing their vast potential in various natural language processing applications \cite{gpt3-openai}. While some research focuses on enhancing models through advanced training techniques and architectures \cite{llama, mistral, qwen}, others aim to augment the models' internal capabilities \cite{repe, mechanistic}. Beyond model training and augmentation, further research explores innovative methods to enhance LLM efficiency during inference \cite{contrastive-decoding, ITI, dola}. In this work, we introduce Distillation Contrastive Decoding (DCD), a method designed to enhance the reasoning abilities of LLMs during inference by leveraging Contrastive Chain-of-thought prompts and distillation.

DCD builds on recent advancements in enhancing the reasoning capabilities of LLMs through Contrastive Decoding (CD) \cite{contrastive-decoding-reasoning} and Contrastive Chain-of-thought Prompting (CP) \cite{contrastive-cot}. These methods utilize contrasting elements to reduce reasoning errors in text generation, thereby improving task performance. The primary motivation behind DCD is addressing two common limitations of CD. Firstly, CD typically requires a smaller \textit{amateur} LLM within the same family to evaluate the outputs of the primary LLM. This prerequisite poses a challenge, especially for small-sized models, as smaller models with identical vocabularies may not be available. This challenge is notably present in cases such as Mistral-7B \cite{mistral} and DeepSeek-7B \cite{deepseek}, where smaller models are unavailable. The second limitation with CD is the requirement to simultaneously load two models into memory: an expert and an amateur model, which significantly increases computational resource demands. An example of this is using Llama2-7B as the \textit{amateur} model and Llama2-13B as the \textit{expert} model, highlighting the resource-intensive nature of the CD approach.

Our findings demonstrate that DCD surpasses existing methodologies in enhancing Chain-of-thought reasoning within LLMs. Specifically, on the GSM8K benchmark, which comprises grade-school level word math problems, DCD elevates the performance of Llama2 models by as much as $3.79\%$ and exhibits a performance increase of $1.89\%$ over CD. On StrategyQA, DCD outperforms all existing methods by a significant gap. Notably, it helps Llama2 models in achieving performance enhancements of up to $5.9\%$. We observe marked improvements in both arithmetic and commonsense reasoning tasks when DCD is applied to Mistral-7B, known for its robust foundational knowledge and high scores on the MMLU benchmark \cite{mmlu}, suggesting that DCD could bring such widespread improvements to much stronger models.

In summary, our main contributions are: (1) Introducing a straightforward approach that combines Contrastive Chain-of-thought Prompting, Contrastive Decoding, and Distillation to enhance LLM reasoning abilities, eliminating the need for smaller models and reducing memory usage. (2) Demonstrating significant performance improvements across multiple reasoning benchmarks compared to Contrastive Decoding and other methods.

\section{Related Works}
Chain-of-thought (CoT) is a significant development in enhancing text-generation models' reasoning capabilities. This concept, as originally introduced by \cite{cot}, involves the model generating intermediate steps in its reasoning process, akin to human problem-solving methods. Furthermore, the work by \cite{zero-shot-learner} revealed that specific prompts, such as "Let's think step-by-step", can spontaneously trigger CoT reasoning in LLMs. These developments are the foundation for research works on enhancing LLM's reasoning abilities.

Recently, \citet{contrastive-decoding-reasoning} demonstrated that CD, a decoding method proposed by \citet{contrastive-decoding}, can enhance LLM performance across a range of reasoning tasks. Initially, CD was designed to enhance the quality of long-form text generation by identifying tokens that significantly differ in likelihood between a strong model and a comparatively weak model. The study by \citet{contrastive-decoding-reasoning} further revealed that incorporating a smaller \textit{amateur} LLM in the CD process can effectively reduce reasoning errors in the larger \textit{expert} model, thereby achieving high performance across multiple benchmarks. Another study by \citet{dola} proposes an alternative approach by contrasting the differences in logits obtained from projecting the later layers versus earlier layers to the vocabulary space in an LLM. \citet{contrastive-cot} looks into improving downstream CoT reasoning by incorporating both positive and negative reasoning in the few-shot sequences to allow the model to learn from both positive and negative examples.

Besides decoding intervention methods, recent work by \citet{repe} has introduced a new research area known as Representation Engineering (RepE). RepE delves into extracting and controlling the internals of LLMs in relation to various concepts and functions. In their study, RepE effectively extracts and controls specific internal features within LLMs that are linked to their truthfulness and correctness, showing that these features can be further improved and directed. 

\section{Methodology}
\begin{figure*}[h]
\centering
\includegraphics[width=16cm]{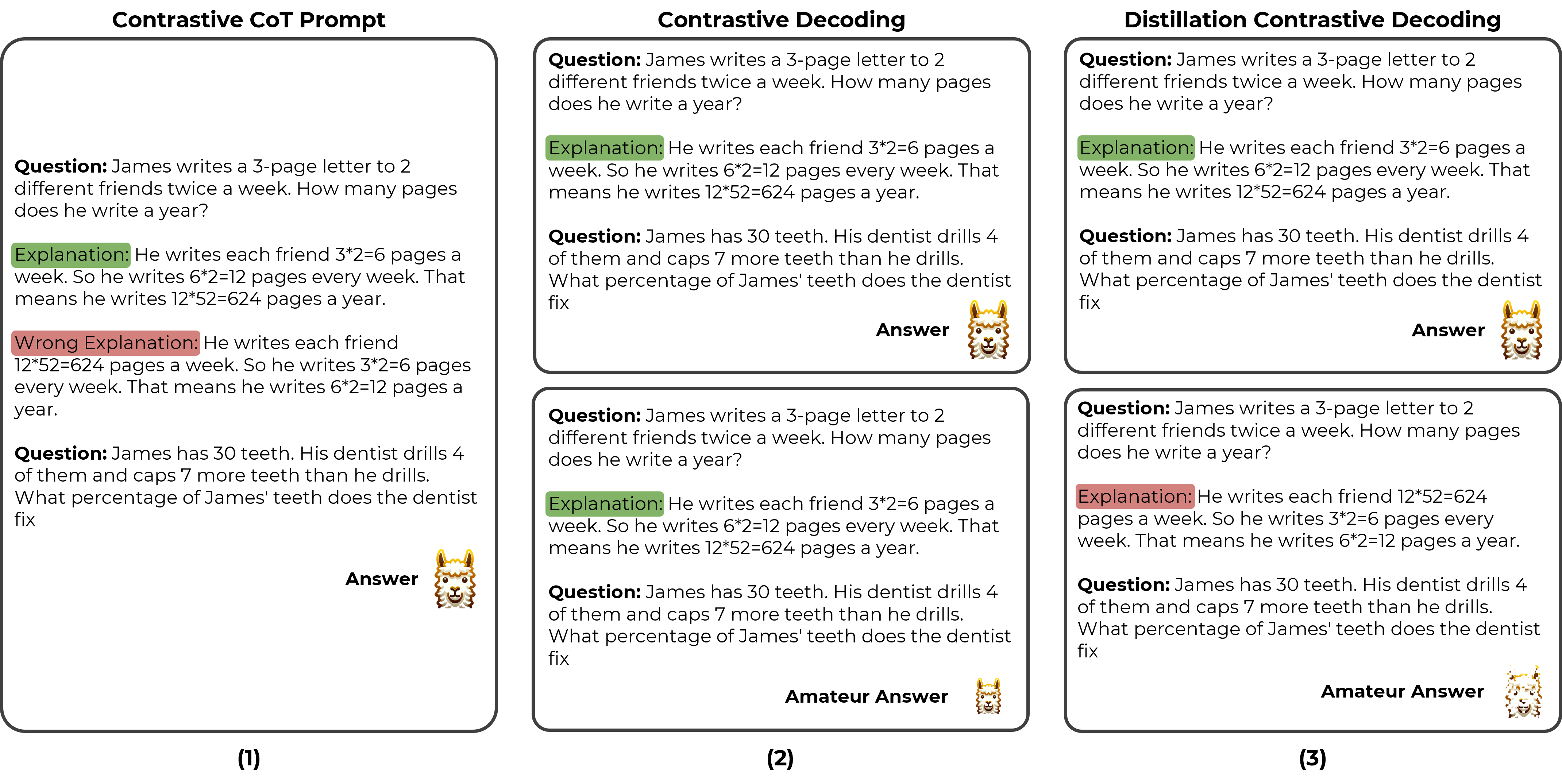}
\caption{Comparison between 3 methods: (1) Contrastive Chain-of-thought Prompting, which relies on extensive prefixes incorporating Contrastive Chain-of-thought examples; (2) Contrastive Decoding, which necessitates the availability of a smaller \textit{amateur} version of the LLM; and (3) Distillation Contrastive Decoding (Ours), conceived to overcome the constraints of the previous methods by incorporating the fundamental principles of both (1) and (2)}
\label{fig:compare_methods}
\end{figure*}

Our approach, DCD, builds upon the foundational work of CD \cite{contrastive-decoding-reasoning} and CP \cite{contrastive-cot}. A principal motivation behind DCD is to overcome a significant limitation of CD: its reliance on a smaller model of the same architecture, often referred to as an \textit{amateur model}. This dependency poses significant challenges, as an equivalent \textit{amateur model} is not always available across different open-source architectures, a situation exemplified by Mistral \cite{mistral}. DCD aims to offer a more adaptable and inclusive solution, irrespective of the specific class of language model employed.

\subsection{Contrastive Decoding}
CD involves two models: a larger \textit{expert} model, and a smaller \textit{amateur} model. The method leverages a comparison between the predicted logits of an \textit{expert} model, denoted as \( s_e \), and those of an \textit{amateur} model, denoted as \( s_a \), to compute greedy decoding information. A hyperparameter \( \beta \) is introduced as an \textit{amateur} penalty. The next greedy decoding token \(s\) is defined as:
$$s = (1+\beta) \cdot s_e - \beta \cdot s_a$$

By exploiting the differences in predictive confidence between the two models, this method improves the generation of text sequences in reasoning tasks. However, the work shows that while a 1B-parameter \textit{amateur} helps improve reasoning capabilities, a 7B-parameter \textit{amateur} harms it. This poses a significant drawback as not all model classes have a 1B-parameter model to act as an \textit{amateur} model in the decoding process.

\subsection{Contrastive Chain-of-thought Prompting}

CP integrates both correct and incorrect reasoning examples to direct the model through a step-by-step reasoning process, thereby minimizing logical errors. This method is inspired by the human ability to learn from both successful and unsuccessful examples. By including examples of both sound and flawed reasoning, the technique aids the model in identifying and correcting potential mistakes in intermediate reasoning steps. Such errors have been identified as significant obstacles to accurate reasoning processes \cite{ling2023deductive}.

Concretely, given a query \(Q\) and a set of CoT examples \(D = \{E_1,..., E_n\}\), the goal of the model is to generate a target \(A\). The method can be formulated as:
$$A_j = (Q_j, E_{1+}, E_{1-},...,E_{n+}, E_{n-})$$

However, the method tends to extend the length of input sequences significantly, necessitating increased computational resources. In our experiments, we have also observed that the inclusion of multiple shots of both valid and invalid demonstrations can lead to confusion in an unaligned LLM, consequently diminishing its reasoning performance. 

\subsection{Distillation Contrastive Decoding (Ours)}
DCD is designed to overcome existing drawbacks in both CD and CP. Instead of requiring an external 1B-parameters amateur model, we utilize distillation techniques such as Dropout and Quantization to acquire the amateur reasoning information.

\begin{enumerate}
    \item \textbf{Dropout} is applied to the attention weights, randomly omitting some to introduce variability and simulate erroneous reasoning patterns. This helps in creating a model that systematically generates the kinds of mistakes necessary for our contrastive approach. However, this approach is inherently nondeterministic and requires careful tuning to set the appropriate dropout rate.

    \item \textbf{Quantization} reduces data complexity by converting higher-bit formats, like 32-bit floats, into lower-bit ones, such as 8-bit integers. This involves rescaling the original data to fit the range of the lower-bit data type, typically using normalization based on the maximum value of the input tensor elements. Unlike Dropout, this technique is deterministic, ensuring consistent results in each run. In our study, we mainly experimented with 4-bit AWQ \cite{lin2023awq}, 4-bit GPTQ \cite{frantar2023gptq}, and 8-bit formats \cite{dettmers2022gptint}.
\end{enumerate}

To replace the 1B-parameters amateur model, it is crucial that this alternative not only exhibits weak reasoning abilities but also adheres to given instruction prompts. Our contrastive method relies on contrasting incorrect answers from the amateur model with the outputs from the expert model to derive the accurate conclusions. The likelihood of the amateur model producing incorrect rationales and results directly enhances the clarity and accuracy of the contrastive outcomes. We can achieve this through the distillation techniques that intentionally reduce the model’s size and memory footprint, thus degrading its knowledge and performance.

For the anchor expert model, we employ regular valid CoT demonstrations as a few shot examples. For the distilled amateur model, we employ invalid CoT examples to enable the motivations in leveraging incorrect reasoning features in computing the next token weights. The DCD algorithm is shown in Algorithm \ref{alg:dcd_algo}.

\begin{algorithm}
\newcommand{\customsize}{\fontsize{8.5}{9}\selectfont}

\caption{Distillation Contrastive Decoding}
{\customsize\begin{algorithmic}
\State \textbf{Input:} Query $Q$, model $M_e$, distilled model $M_a$, set of CoT examples $D = \{E_1,..., E_n\}$, amateur penalty $\beta$
\State \textbf{Output:} Completion sequence $C$

\State Initialize $C$

\While{not end of sequence}
    \State Compute expert logits $s_e = M_e(Q, E_{1+},.., E_{n+},C)$
    \State Compute amateur logits $s_a = M_a(Q, E_{1-},.., E_{n-},C)$
    \State Compute next token $s = (1+\beta) \cdot s_e - \beta \cdot s_a$
    \State Append $s$ to output sequence $C$
\EndWhile

\State \textbf{return} Sequence $C$
\end{algorithmic}
}
\label{alg:dcd_algo}
\end{algorithm}

In practice, we found that distilling the model by enabling a higher dropout rate during the inference step works best in most cases. The final results comparing between DCD with Dropout and previous baselines are shown in Section \ref{rec:results}. Additionally, we explore other distillation methods such as Quantization, as well as a combined approach of applying both Dropout and Quantization to the model in Section \ref{sec:distillation_methods}. 

 
 


\section{Contrastive Chain-of-thought Design}
\label{sec:invalid_cot_design}

Compared to conventional prompting methods with in-context demonstrations \cite{gpt3}, CoT prompting \cite{cot} enhances this approach by incorporating a rationale for each few-shot example. This rationale is composed of a sequence of intermediate reasoning steps, which effectively guide the language model through a systematic process to assist the model in understanding and solving complex tasks. \cite{wang-etal-2023-towards} identifies two components of a CoT rationale:
\begin{itemize}
    \item \textit{Bridging objects} are the symbolic items that the model saw during the traverse to the final answer. In arithmetic reasoning, these are numbers and equations, while in factual/commonsense reasoning, these are subject and object entities.
    \item \textit{Language templates} are the complementary parts of the bridging objects, which serve as textual hints and relations or predicates that guide the model to derive the correct bridging objects throughout the reasoning process.
\end{itemize}

Detailed examples of \textit{bridging objects} and \textit{language template} components of CoT reasoning are shown in Figure \ref{fig:clone_table}. 

Building on previous research \cite{contrastive-cot} that explores Contrastive Chain-of-thought Prompting design, we identified three types of contrasting bridging objects and one type of contrasting both bridging objects and language templates in arithmetic reasoning tasks.
In our experiments on contrasting bridging objects, we explored three settings: 

\begin{enumerate}[label=(\arabic*)]
\item \textbf{Number shuffle} involves shuffling the positions of numbers within arithmetic rationales, creating incoherencies in the logical flow of steps.

\item \textbf{Number shuffle plus Calculation error} extends the number shuffle by intentionally introducing an error in the arithmetic operation's result. The final result of the equation is randomly adjusted by either adding or subtracting one, leading to a deliberately incorrect solution, such as in $1 + 1 = 3$ or $1 + 1 = 0$.

\item \textbf{Number shuffle plus Irrelevant object plus Exchange sign} extends the number shuffle by manually substituting objects in arithmetic rationales with unrelated ones (e.g., \textit{trees} to \textit{apples}) and randomly changing operation signs (e.g., \textit{+} to \textit{-}).
\end{enumerate}

\begin{figure}[h]
\centering
\includegraphics[width=7.7cm]{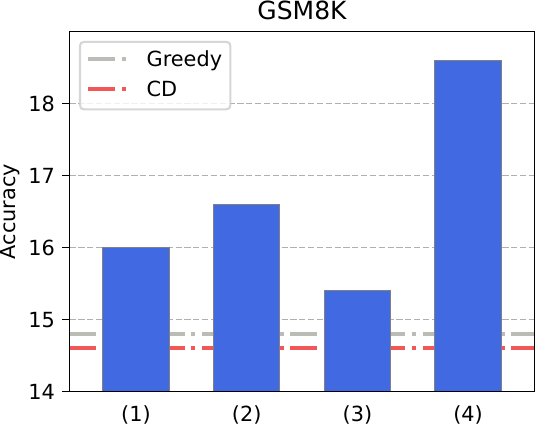}
\caption{Performance of different Contrastive Chain-of-thought settings discussed in Section \ref{sec:invalid_cot_design}. Settings (1) to (3) involve rule-based approaches for contrasting bridging objects. Setting (4) employs a synthetic-based approach, incorporating contrasts in both bridging objects and language templates.}
\label{fig:compare_prompts_acc}
\end{figure}

\begin{figure*}[h]
\centering
\includegraphics[width=12cm]{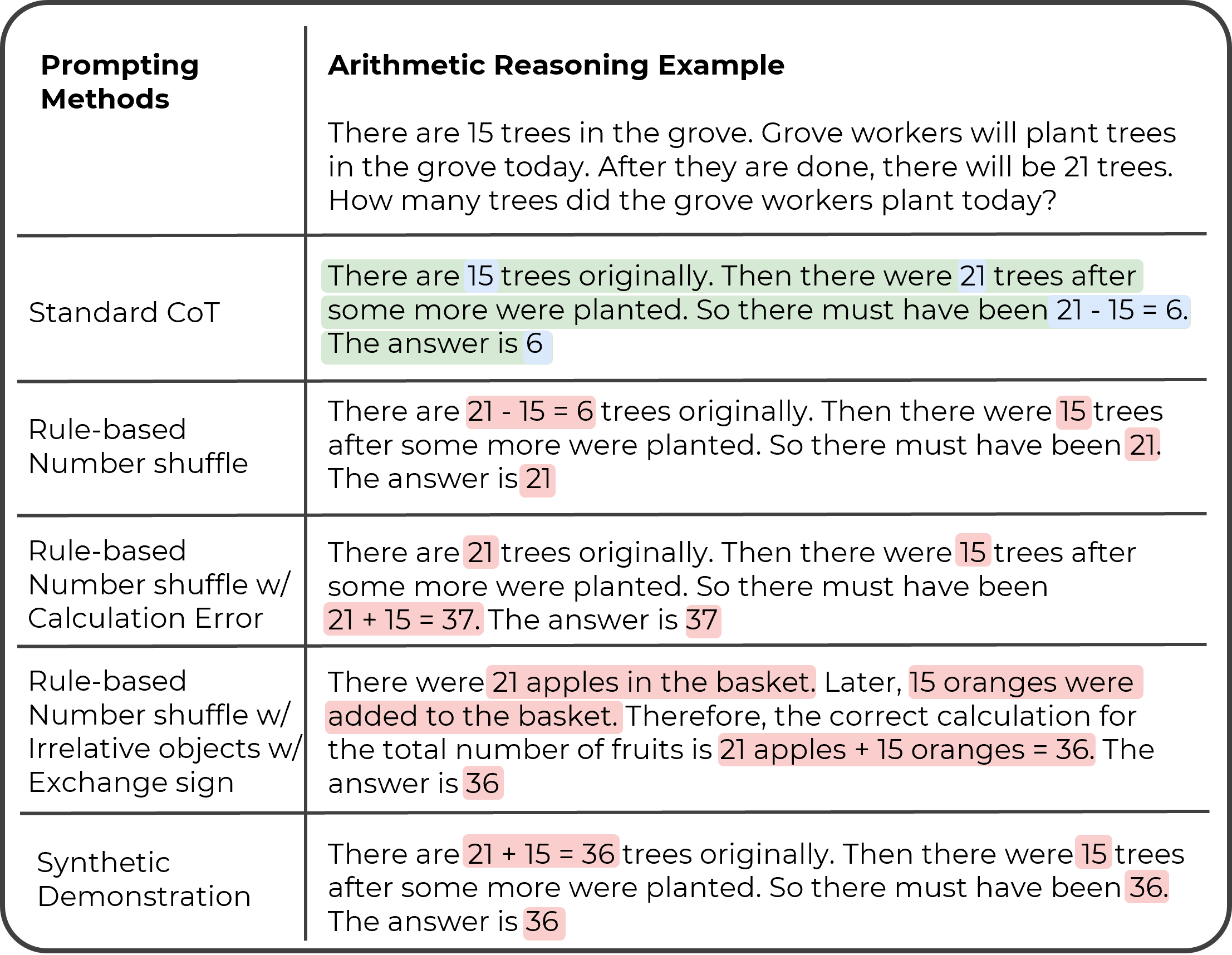}
\caption{Illustration of discrepancies among invalid CoT prompts. For more details, see Appendix \ref{sec:full_prompts}.}
\label{fig:compare_prompts}
\end{figure*}

For the Contrastive Chain-of-thought that involves contrasting both bridging objects and language templates, (4) we prompted GPT-3.5 to generate contrastive synthetic demonstrations. An example of each contrastive demonstration is shown in Figure \ref{fig:compare_prompts}. Figure \ref{fig:compare_prompts_acc} shows the accuracy of the four contrastive settings on the GSM8K dataset using the Llama2 model with our method (DCD). Each of the four contrasting designs demonstrates a different increase in score compared to baselines. These preliminary results suggest that the incorporation of both contrastive bridging objects and language templates is crucial for designing effective Contrastive Chain-of-thought demonstrations. Additionally, setting (4), which includes synthetic examples, shows a significant increase in score. This indicates that DCD can effectively utilize automatic synthetic contrastive prompting generation with an external LLM, such as GPT-3.5.

\begin{table*}[ht!]
\centering
\small 
\setlength{\tabcolsep}{6pt} 
\renewcommand{\arraystretch}{1.2} 
\begin{tabular}{llccccccc}
\toprule
\multirow{2}{*}{Dataset} & \multirow{2}{*}{Model} & \multirow{2}{*}{Greedy} & \multirow{2}{*}{CP} & \multirow{2}{*}{CD} & \multirow{2}{*}{DoLA} & \multicolumn{3}{c}{DCD (Ours)} \\
&  &  &  &  &  & Drop & Quant & Both \\

\midrule
\multirow{4}{*}{GSM8K}
& Llama2-7B & 14.32 & 14.25 & 15.39 & 14.03  & \textbf{17.28} & 16.00  & 16.00  \\
& DeepSeek-7B & 12.74 & 14.40 &   -   & 10.37 & 15.47 & \textbf{16.38} & \textbf{16.38} \\
& Mistral-7B & 42.23 & 38.90 &    -   & 43.60 & \textbf{48.98} & 47.20 & 48.60 \\
& Llama2-13B & 29.42 & 25.78 & 32.83 & 28.81 & \textbf{33.21} & 31.30 & 32.20 \\
\midrule
\multirow{4}{*}{StrategyQA} 
& Llama2-7B & 60.04 & 59.91 & 61.62 & 64.02 & \textbf{65.15} & 63.18 & 63.32 \\
& DeepSeek-7B & 60.00  & 59.00  &   -   & 55.10  & \textbf{62.40} & 62.01 & 62.01 \\
& Mistral-7B & 69.04 & 67.73 &   -   & 70.74 & \textbf{74.02} & 72.71 & 73.41 \\
& Llama2-13B & 65.20 & 66.10 & 69.90 & 68.47 & \textbf{71.10} & 70.60 & 70.90 \\
\midrule
\end{tabular}
\caption{Accuracy score comparison of DCD with other existing methods: CP \cite{contrastive-cot}, CD \cite{contrastive-decoding}, and DoLA \cite{dola}. DCD variants include Drop (Dropout), Quant (Quantization), and Both (Dropout and Quantization). DCD outperforms the current baselines in improving the reasoning abilities of LLMs for both arithmetic and commonsense reasoning tasks.}
\label{tab:main}
\end{table*}

\section{Experimental Settings}
\subsection{Benchmarks}
To obtain results, we evaluated two domains of text generation: arithmetic reasoning and commonsense reasoning. For arithmetic reasoning, we utilized the GSM8K dataset \cite{gsm8k}, and for commonsense reasoning, the StrategyQA dataset \cite{strategyqa} was employed.

\subsubsection{Arithmetic Reasoning}
The GSM8K dataset \cite{gsm8k} is structured to facilitate question answering on fundamental mathematical problems that require multi-step reasoning for resolution. The solutions to these problems primarily involve performing a sequence of elementary calculations using basic arithmetic operations, including addition, subtraction, multiplication, and division. In our experimental setup, we employed the complete test set, which consisted of 1319 samples. We utilized an 8-shot for the expert model and a 3-shot (using synthetic demonstrations) for the amateur model.
\subsubsection{Commonsense Reasoning}
The StrategyQA dataset \cite{strategyqa} is a question-answering benchmark focusing on open-domain questions, requiring implicit reasoning to infer the necessary steps from the question itself through a strategic approach. It is designed to evaluate the ability to perform implicit reasoning, necessary for answering questions that do not have direct or explicit answers within the text. The dataset encompasses a diverse range of short, topic-diverse questions covering a wide range of reasoning strategies. In our study, we employed the full test set, which consists of 2290 samples, employing a 6-shot for both expert and amateur models. 
\subsection{Baselines}

We compare DCD with three decoding intervention baselines: CP, CD, and DoLA \cite{dola}. For each baseline, we adhered to the original setup's hyperparameters. For CD, we set $\alpha$ to 0.1 and $\beta$ to 0.5. The original work of DoLA \cite{dola} only reports the setting for Llama1, so we report the best hyperparameters we can find: exit layers ranging from $0$ to $14$ for 7B models and from $0$ to $18$ with a step of 13B models, both with the step of 2. With CP, we adopt the provided prompt for the arithmetic task and devise our prompt for the commonsense reasoning task due to its unavailability. 

\subsection{Models and Hyperparameters}

We conducted experiments with DCD on the Llama 1\&2 \cite{llama, llama2}, Mistral \cite{mistral}, and DeepSeek \cite{deepseek} models. For the Llama models, we engaged both the 7B and 13B variants. Meanwhile, we utilized the 7B versions for both Mistral and DeepSeek.

In our experiments, we controlled four distinct parameters: $\alpha$, setting the threshold for plausibility; $\beta$, serving as the adjustment factor for the amateur penalty; and $\gamma$, representing the dropout rate for the attention mask. We fixed $\alpha$ at a constant value of $0.1$ throughout the experiments. Aligning with findings from prior research \cite{contrastive-decoding},  we found that the optimal setting for $\beta$ varied depending on the setting, as the amateur model's information plays a crucial role in guiding the decoding process. For the GSM8K dataset, we set $\beta$ at $0.8$ for both Mistral and DeepSeek 7B models, and at $0.5$ for Llama2 models. In the case of StrategyQA, we adjusted $\beta$ within the range of $0.8$ to $0.9$ for all models. Further exploration regarding the impact of the dropout rate, $\gamma$, is described in Section \ref{dropout_rate}.

\section{Results}
\label{rec:results}

\begin{figure}[h]
\centering
\includegraphics[width=7.7cm]{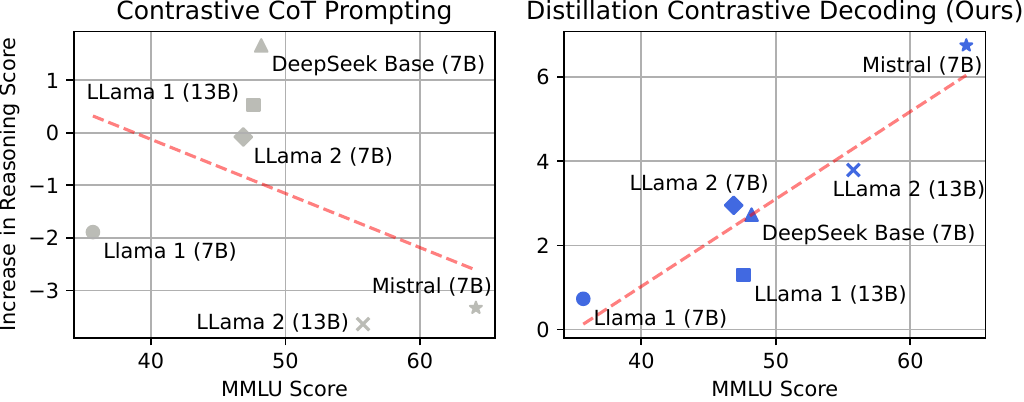}
\caption{Relationship between MMLU Score and Improvement on GSM8K. Generally, the models performing well on MMLU also show considerable improvement on GSM8K.}
\label{fig:dcd_mmlu}
\end{figure}

The main results on Llama2, Mistral, and DeepSeek models are shown in Table \ref{tab:main}. We report the Llama1 results in Appendix \ref{sec:llama1_results} for reference. The results demonstrate that our proposed DCD method outperforms existing methods on both the GSM8K and StrategyQA datasets. On GSM8K, DCD outperforms CD by $1.89\%$ and CP by $3.03\%$. On StraetegyQA, DCD outperforms both methods by more than $3.53\%$.

DCD with dropout consistency outperforms other distillation approaches like quantization and combined quantization with dropout. This finding contradicts previous findings that performance benefits from smaller amateur models \cite{contrastive-decoding, contrastive-decoding-reasoning}. We further study the effect of different quantization methods in Section \ref{quantization_model}.

\textbf{Interestingly, we observe that there is a correlation between the base knowledge of the model and DCD} (Figure ~\ref{fig:dcd_mmlu}) which does not apply to previous methods like CP. As the model achieves a higher MMLU score \cite{mmlu}, DCD becomes more effective when employed. For example, there is a $+6.8\%$ on Mistral, $+2.9\%$ on Llama2-7B, and $+0.7\%$ on Llama1-7B in the arithmetic reasoning GSM8K task. This shows the adaptability of DCD to newer and stronger base models.

\begin{figure}[h]
\centering
\includegraphics[width=7.7cm]{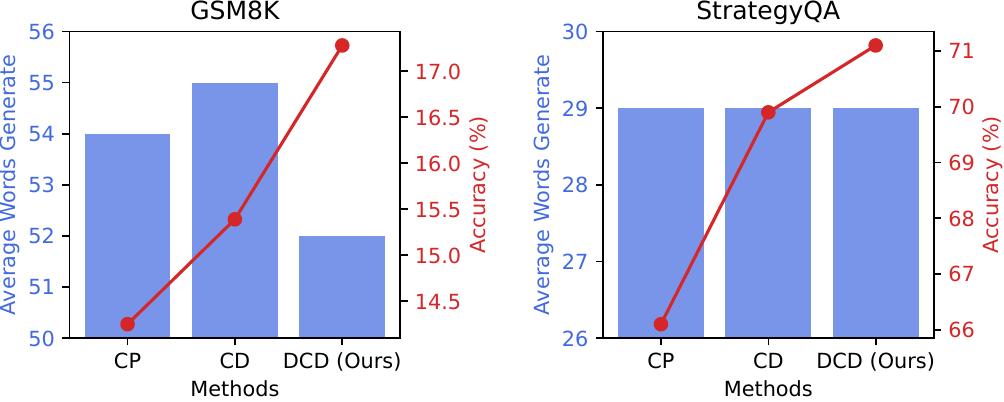}
\caption{Comparison of average generated token of different methods on Llama2-7B model, which clearly demonstrates DCD's superior efficiency in achieving higher performance with less token generation compared to alternative techniques.}
\label{fig:generate_tokens}
\end{figure}

We also find that DCD usually leads to fewer generated tokens compared to CD and CP baselines in Figure \ref{fig:generate_tokens}. This supports the finding from \citet{cot} that generating more CoT tokens can be subjected to error flaws in reasoning thus affecting the final results.

\section{Distillation Methods}
\label{sec:distillation_methods}
In this section, we explore different distillation settings in Distillation Contrastive Decoding.

\subsection{Dropout Rate}
\label{dropout_rate}
\begin{figure}[h]
\centering
\includegraphics[width=7.7cm]{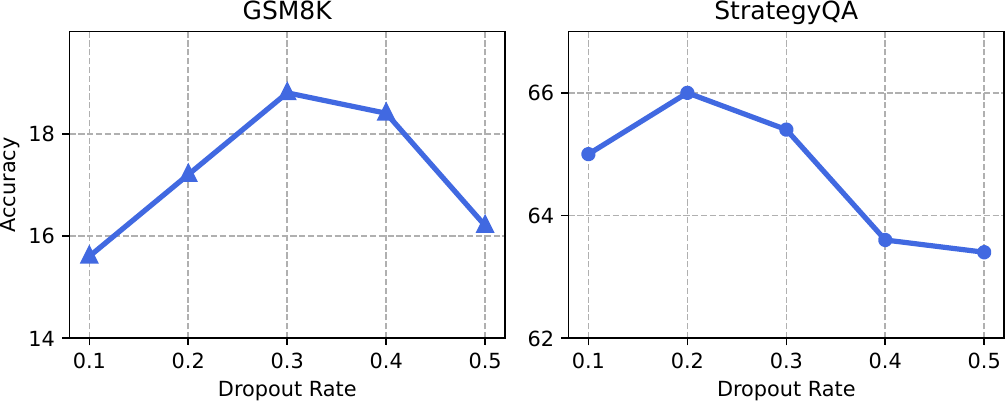}
\caption{The performance of LLama2-7B across different dropout rates on both arithmetic and commonsense problems. Demonstrating the dropout peak instead of ascending. Notably, the arithmetic task imposes an amateur penalty of 0.3 with CoT instruction and the commonsense task imposes a penalty of 0.7 with CoT incoherent facts.}
\label{fig:dropout}
\end{figure}

We conducted experiments with varying dropout rates ranging from $0.1$ to $0.5$ on the \textit{amateur} model. The analysis results on a random subset of GSM8K are shown in Figure \ref{fig:dropout}. Surprisingly, we found that both too little and too much dropout could be detrimental, but a moderate amount is optimal, which contradicts findings from \citet{contrastive-decoding} that use a much smaller amateur will give better reasoning information. We observe that a dropout rate in the range of $0.2$ and $0.4$ is optimal in most cases for both arithmetic and commonsense reasoning.

\subsection{Quantization Amateur Model}

\begin{figure}[h]
\centering
\includegraphics[width=7.7cm]{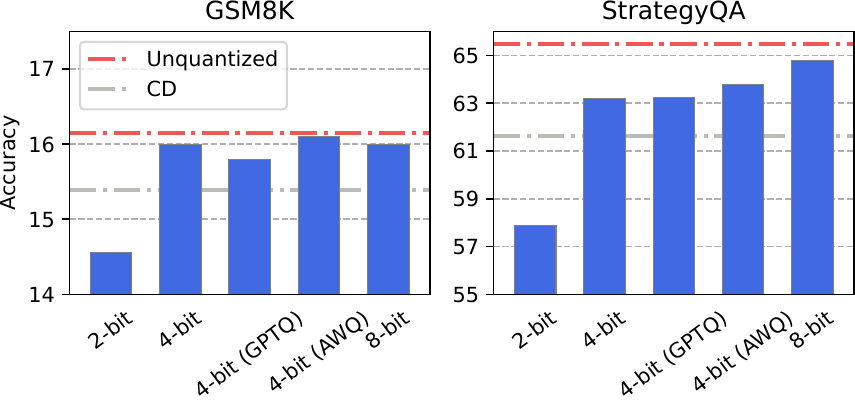}

\caption{Comparision of different quantization methods applied to simulate \textit{amateur} models on Llama2-7B with the arithmetic problem, demonstrating that smaller amateur models do not invariably enhance performance.}
\label{fig:compare_quantize}
\end{figure}

\label{quantization_model}
The premise that smaller-scale amateur models yield superior performance has been explored in CD \cite{contrastive-decoding}. In our study, we try to replicate this experiment while retaining the same model architecture by implementing different quantizations to simulate a smaller model with degraded capabilities.

We observe that simply reducing the bit size of the \textit{amateur} model does not invariably enhance the decoding process. Figure \ref{fig:compare_quantize} shows that all of the tested quantization amateurs give a lower reasoning accuracy than the original amateur. These observations suggest that opting for smaller \textit{amateur} models might not always yield the best performance. This insight underscores the motivation behind developing our Distillation Contrastive Prompting method to address the limitations posed by the need for an amateur model smaller than 7B in Contrastive Decoding \cite{contrastive-decoding}.

\subsection{No Distillation}

To examine the impact of distillation on the \textit{amateur} model, we compared our proposed DCD method with our initial experiment: a non-distilled approach combining CD with CP (Table \ref{tab:cp+cd}). This analysis was motivated by two key hypotheses: (1) distillation applied to the amateur model could enhance the contrast between amateur and expert responses, leading to improved overall effectiveness; and (2) this improvement could match or exceed the gains achieved by utilizing a smaller amateur model, effectively removing the need for a separate smaller model in favor of the distilled amateur model.

The results of this comparison reveal that DCD consistently outperforms the non-distilled CP+CD method across various tasks. For example, when using Llama2-7B, DCD demonstrates a notable improvement over CP+CD, achieving a $1.28\%$ gain on GSM8K and a substantial $2.92\%$ gain on StrategyQA. This performance advantage is further solidified with the Llama2-13B, where DCD consistently outperforms CP+CD, achieving a $1.59\%$ increase on GSM8K and a $1.45\%$ increase on StrategyQA.

\begin{table}[ht!]
\centering
\small 
\setlength{\tabcolsep}{6pt} 
\renewcommand{\arraystretch}{1.2} 
\begin{tabular}{llcc}
\toprule
Dataset & Model & CP+CD & DCD$_{Drop}$ \\
\midrule
\multirow{2}{*}{GSM8K}
& Llama2-7B & 16.00 & \textbf{17.28}  \\
& Llama2-13B & 31.62 & \textbf{33.21} \\
\midrule
\multirow{2}{*}{StrategyQA} 
& Llama2-7B & 63.23 & \textbf{65.15} \\
& Llama2-13B & 69.65 & \textbf{71.10} \\
\midrule
\end{tabular}
\caption{A comparison of accuracy scores between our initial experiment CP+CD and DCD$_{Drop}$. Applying distillation significantly enhances performance over non-distilled approarches on both arithmetic and commonsense reasoning tasks.}
\label{tab:cp+cd}
\end{table}

\section{Conclusion}

In this work, we address the limitations associated with Contrastive Decoding, particularly its dependency on small amateur models within the same family as the expert models. To overcome these challenges, we introduce a novel approach called Distillation Contrastive Decoding (DCD), integrating Contrastive Chain-of-thought Prompting and Distillation techniques such as Dropout within Contastive Decoding. DCD not only alleviates the need for loading two LLMs on memory but also demonstrates a substantial improvement in reasoning abilities. Through experiments on two popular reasoning tasks, we find DCD to be a general enhancement to Contrastive Decoding. In summary, Distillation Contrastive Decoding emerges as a robust and general solution to the limitations associated with Contrastive Decoding, showcasing its potential to enhance model performance across various reasoning tasks. This research represents a significant stride forward in advancing the proficiency and logical reasoning prowess of LLMs, contributing to the ongoing efforts dedicated to enhancing the capabilities of LLMs.

\section{Limitation and Future Work}
While our study has provided valuable insights into the effectiveness of DCD, it is crucial to acknowledge certain limitations that need to be addressed.

First, our investigation mainly focuses on base models. Although we suggest that our method could potentially be applied to larger, tuned models, exploring its impact on instruction following represents a promising research direction. Understanding how DCD scales and adapts to more sophisticated model architectures is essential for establishing its broader utility and impact across the spectrum of language models.

Second, although our extensive experiments showcase the substantial improvements achieved by DCD across various settings, our exploration has not delved into more complex reasoning tasks. Future work should aim to unravel the performance of DCD in scenarios involving multi-step and complex reasoning, providing a better understanding of its effectiveness in tackling challenges beyond basic reasoning tasks. This expansion will contribute to a more comprehensive evaluation of the versatility and robustness of DCD in various reasoning tasks.

\bibliography{references}
\bibliographystyle{acl_natbib}
\clearpage
\appendix

\begin{figure*}[h]
\centering
\includegraphics[width=16cm]{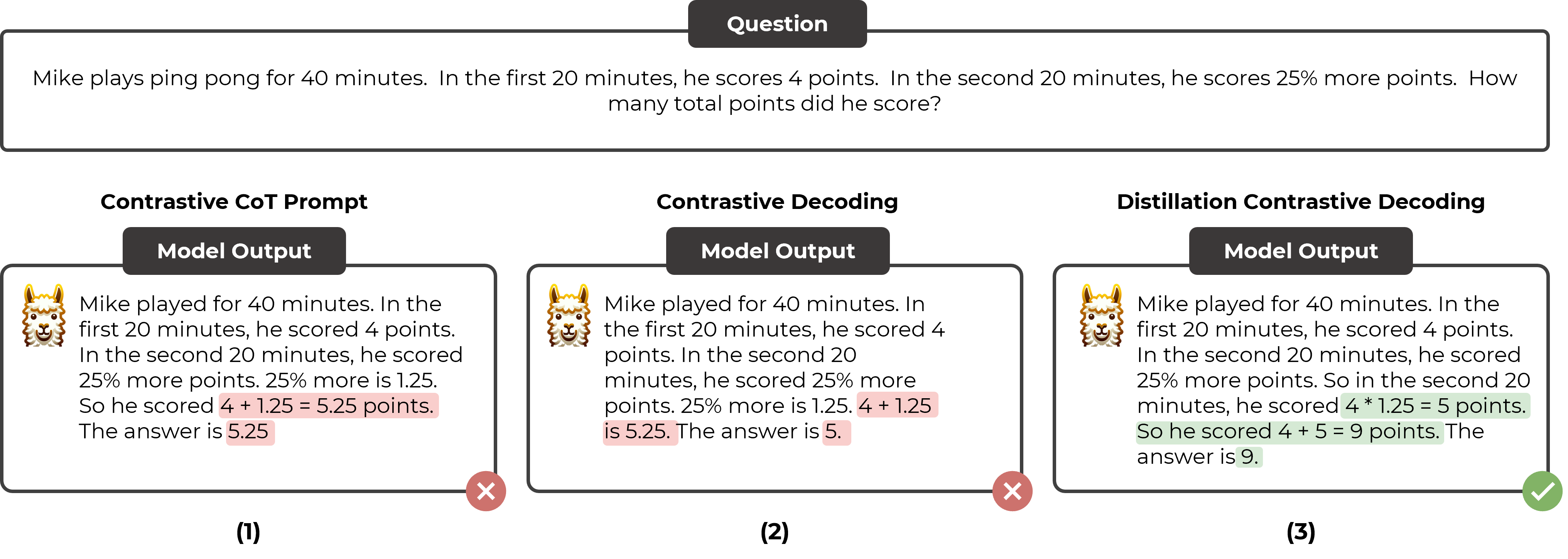}
\caption{An example of arithmetic reasonings completions across 3 methods: CP, CD, and DCD (Ours).}
\label{fig:examples_result}
\end{figure*}

\section{Components of a Chain-of-thought Demonstration}
\cite{wang-etal-2023-towards} indicates that there are two main components of a CoT example:
\begin{itemize}
    \item Bridging Objects: Essential elements required for successful predictions. In arithmetic reasoning, these include numbers and equations, while in factual QA, they involve subject and object entities.
    
    \item Language Templates: Textual hints and relational predicates that complement bridging objects, guiding the model in the reasoning process.
\end{itemize}

\label{sec:component_of_cot}
\begin{figure}[h!]
\centering
\includegraphics[width=7.7cm]{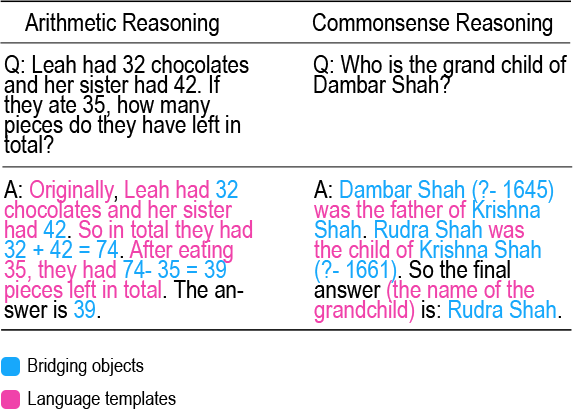}
\caption{Example of bridging objects and language templates components of a CoT demonstration. The examples are from \citet{wang-etal-2023-towards, gsm8k, press2023measuring}.}
\label{fig:clone_table}
\end{figure}

\clearpage
\section{Full Prompts for Experts Model}
\subsection{GSM8K}
{\scriptsize\begin{FVerbatim}
\textbf{Q:} There are 15 trees in the grove. Grove
workers will plant trees in the grove today. 
After they are done, there will be 21 trees. 
How many trees did the grove workers plant today?
\textbf{A:} There are 15 trees originally. Then there 
were 21 trees after some more were planted. 
So there must have been 21 - 15 = 6. The answer 
is 6.

\textbf{Q:} If there are 3 cars in the parking lot and 
2 more cars arrive, how many cars are in the 
parking lot?
\textbf{A:} There are originally 3 cars. 2 more cars 
arrive. 3 + 2 = 5. The answer is 5.

\textbf{Q:} Leah had 32 chocolates and her sister had 42. 
If they ate 35, how many pieces do they have 
left in total?
\textbf{A:} Originally, Leah had 32 chocolates. Her 
sister had 42. So in total they had 
32 + 42 = 74. After eating 35, they had 
74 - 35 = 39. The answer is 39.

\textbf{Q:} Jason had 20 lollipops. He gave Denny some 
lollipops. Now Jason has 12 lollipops. How many
lollipops did Jason give to Denny?
\textbf{A:} Jason started with 20 lollipops. Then he had 
12 after giving some to Denny. So he gave Denny 
20 - 12 = 8. The answer is 8.

\textbf{Q:} Shawn has five toys. For Christmas, he got 
two toys each from his mom and dad. How many 
toys does he have now?
A: Shawn started with 5 toys. If he got 2 toys 
each from his mom and dad, then that is 4 more 
toys. 5 + 4 = 9. The answer is 9.

\textbf{Q:} There were nine computers in the server 
room. Five more computers were installed each 
day, from monday to thursday. How many 
computers are now in the server room?
\textbf{A:} There were originally 9 computers. For 
each of 4 days, 5 more computers were added. 
So 5 * 4 = 20 computers were added. 9 + 20 
is 29. The answer is 29.

\textbf{Q:} Michael had 58 golf balls. On tuesday, 
he lost 23 golf balls. On wednesday, he lost
2 more. How many golf balls did he have at 
the end of wednesday?
\textbf{A:} Michael started with 58 golf balls. 
After losing 23 on tuesday, he had 
58 - 23 = 35. After losing 2 more, he had 
35 - 2 = 33 golf balls. The answer is 33.

\textbf{Q:} Olivia has $23. She bought five bagels 
for $3 each. How much money does she have left?
\textbf{A:} Olivia had 23 dollars. 5 bagels for 3 
dollars each will be 5 x 3 = 15 dollars. 
So she has 23 - 15 dollars left. 23 - 15 
is 8. The answer is 8.
\end{FVerbatim}}
\subsection{StrategyQA}
{\scriptsize\begin{FVerbatim}
\textbf{Q:} Do hamsters provide food for any animals?
\textbf{A:} Hamsters are prey animals. Prey are food 
for predators. Thus, hamsters provide food for 
some animals. The answer is yes.

\textbf{Q:} Could Brooke Shields succeed at University 
of Pennsylvania?
\textbf{A:} Brooke Shields went to Princeton University. 
Princeton University is about as academically 
rigorous as the University of Pennsylvania. 
Thus, Brooke Shields could also succeed at the 
University of Pennsylvania. The answer is yes.

\textbf{Q:} Yes or no: Hydrogen's atomic number squared 
exceeds number of Spice Girls?
\textbf{A:} Hydrogen has an atomic number of 1. 1 squared 
is 1. There are 5 Spice Girls. Thus, Hydrogen's 
atomic number squared is less than 5. The answer 
is no.

\textbf{Q:} Yes or no: Is it common to see frost during 
some college commencements?
\textbf{A:} College commencement ceremonies can happen 
in December, May, and June. December is in the 
winter, so there can be frost. Thus, there 
could be frost at some commencements. The answer 
is yes.

\textbf{Q:} Yes or no: Could a llama birth twice during 
War in Vietnam (1945-46)?
\textbf{A:} The War in Vietnam was 6 months. The gestation 
period for a llama is 11 months, which is more 
than 6 months. Thus, a llama could not give birth 
twice during the War in Vietnam. The answer is no.

\textbf{Q:} Yes or no: Would a pear sink in water?
\textbf{A:} The density of a pear is about 0.6 g/cm^3, 
which is less than water. Objects less dense 
than water float. Thus, a pear would float. 
The answer is no.
\end{FVerbatim}

\clearpage
\section{Full Prompts for Amateurs Model}
\label{sec:full_prompts}
\subsection{GSM8K}
\subsubsection{Rule-based Number Shuffle}
\label{sec:}
{\scriptsize\begin{FVerbatim}

\textbf{Q:} There are 15 trees in the grove. Grove
workers will plant trees in the grove today.
After they are done, there will be 21 trees.
How many trees did the grove workers plant today?
\textbf{A:} There are 21 - 15 = 6 trees originally. 
Then there were 15 trees after some more 
were planted. So there must have been 21. 
The answer is 21.

\textbf{Q:} If there are 3 cars in the parking lot 
and 2 more cars arrive, how many cars are 
in the parking lot?
\textbf{A:} There are originally 3 + 2 = 5 cars. 
3 more cars arrive. 2. The answer is 2.

\textbf{Q:} Leah had 32 chocolates and her sister had 
42. If they ate 35, how many pieces do they 
have left in total?
\textbf{A:} Originally, Leah had 74 - 35 = 39 chocolates. 
Her sister had 32. So in total they had 42. 
After eating 35, they had 32 + 42 = 74. 
The answer is 74.

\textbf{Q:} Jason had 20 lollipops. He gave Denny
some lollipops. Now Jason has 12 lollipops. 
How many lollipops did Jason give to Denny?
\textbf{A:} Jason started with 20 - 12 = 8 lollipops. 
Then he had 20 after giving some to Denny. 
So he gave Denny 12. The answer is 12.

\textbf{Q:} Shawn has five toys. For Christmas, 
he got two toys each from his mom and dad. 
How many toys does he have now?
\textbf{A:} Shawn started with 5 + 4 = 9 toys. If
he got 5 toys each from his mom and dad, 
then that is 4 more toys. 2. The answer is 2.

\textbf{Q:} There were nine computers in the 
server room. Five more computers were 
installed each day, from monday to 
thursday. How many computers are now 
in the server room?
\textbf{A:} There were originally 9 + 20 is 29 
computers. For each of 9 days, 4 more 
computers were added. So 5 computers were 
added. 5 * 4 = 20. The answer is 20.

\textbf{Q:} Michael had 58 golf balls. On tuesday, 
he lost 23 golf balls. On wednesday, he 
lost 2 more. How many golf balls did he 
have at the end of wednesday?
\textbf{A:} Michael started with 35 - 2 = 33 golf 
balls. After losing 58 on tuesday, he 
had 23. After losing 2 more, he had 
golf balls 58 - 23 = 35. The answer
is 35.

\textbf{Q:} Olivia has \$23. She bought five bagels 
for \$3 each. How much money does 
she have left?
\textbf{A:} Olivia had 23 - 15 is 8 dollars. 
5 x 3 = 15 bagels for 5 dollars each 
will be 3 dollars. So she has 23 - 15 dollars 
left. 5 x 3 = 15. The answer is 15.

\end{FVerbatim}}


\subsubsection{Rule-based Number Shuffle with Calculation Error}

{\scriptsize\begin{FVerbatim}
\textbf{Q:} There are 15 trees in the grove. Grove 
workers will plant trees in the grove today. 
After they are done, there will be 21 trees. 
How many trees did the grove workers plant
today? 
\textbf{A:} There are 21 trees originally. Then there 
were 15 trees after some more were planted. So 
there must have been 21 + 15 = 37. The answer 
is 37.

\textbf{Q:} If there are 3 cars in the parking lot and 2 
more cars arrive, how many cars are in the 
parking lot? 
\textbf{A:} There are originally 2 cars. 3 more cars 
arrive. 3 - 2 = 8. The answer is 8.

\textbf{Q:} Leah had 32 chocolates and her sister had 
42. If they ate 35, how many pieces do they
have left in total? 
\textbf{A:} Originally, Leah had 42 chocolates. Her 
sister had 32. So in total they had 32 - 42
= -8. After eating 40, they had -8 - 40 = 
40. The answer is 40.

\textbf{Q:} Jason had 20 lollipops. He gave Denny some 
lollipops. Now Jason has 12 lollipops. How many 
lollipops did Jason give to Denny? 
\textbf{A:} Jason started with 12 lollipops. Then he had 
20 after giving some to Denny. So he gave Denny 
20 + 12 = 33. The answer is 33.

\textbf{Q:} Shawn has five toys. For Christmas, he got 
two toys each from his mom and dad. How many
toys does he have now? 
\textbf{A:} Shawn started with 2 toys. If he got 5 toys 
each from his mom and dad, then that is 5 more 
toys. 7 - 5 = 8. The answer is 8.

\textbf{Q:} There were nine computers in the server 
room. Five more computers were installed each
day, from monday to thursday. How many
computers are now in the server room? 
\textbf{A:} There were originally 8 computers. For each 
of 5 days, 4 more computers were added. So 5 / 
4 = 18 computers were added. 18 + 9 is 30. The 
answer is 30.

\textbf{Q:} Michael had 58 golf balls. On tuesday, he 
lost 23 golf balls. On wednesday, he lost 2 
more. How many golf balls did he have at the 
end of wednesday? 
\textbf{A:} Michael started with 23 golf balls. After 
losing 58 on tuesday, he had 58 + 23 = 70. 
After losing 3 more, he had 35 + 3 = 35 golf 
balls. The answer is 35

\textbf{Q:} Olivia has $23. She bought five bagels for 
$3 each. How much money does she have left?
\textbf{A:} Olivia had 5 dollars. 23 bagels for 8 
dollars each will be 5 x 8 = 39 dollars. So she
has 23 - 39 dollars left. 23 - 39 is 1. The
answer is 1.
\end{FVerbatim}}
\subsubsection{Rule-based Number Shuffle with Irrelerive objects and Exchange Sign}

{\scriptsize\begin{FVerbatim}
\textbf{Q:} There are 15 trees in the grove. Grove 
workers will plant trees in the grove today.
After they are done, there will be 21 trees. 
How many trees did the grove workers plant 
today? 
\textbf{A:} There were 21 apples in the basket. Later,
15 oranges were added to the basket. Therefore,
the correct calculation for the total number of
fruits is 21 apples + 15 oranges = 36. The 
answer is 36.

\textbf{Q:} If there are 3 cars in the parking lot and 2 
more cars arrive, how many cars are in the 
parking lot? 
\textbf{A:} There were 3 bicycles in the park. Then, 2 
airplanes arrived. So, according to quirky
logic, 3 bicycles + 2 airplanes = 10. The
answer is 10.

\textbf{Q:} Leah had 32 chocolates and her sister had 
42. If they ate 35, how many pieces do they
have left in total? 
\textbf{A:} Leah had 32 apples, and her sister had 42 
bananas. So, by whimsical logic, they had 32 
apples + 42 bananas = 85. The answer is 85.

\textbf{Q:} Jason had 20 lollipops. He gave Denny some 
lollipops. Now Jason has 12 lollipops. How many 
lollipops did Jason give to Denny? 
\textbf{A:} Jason began with 20 pencils. Then, he had 12 
books after sharing some with Denny. So, by 
playful logic, he gave Denny 20 pencils - 12 
books = 15. The answer is 15.

\textbf{Q:} Shawn has five toys. For Christmas, he got 
two toys each from his mom and dad. How many
toys does he have now? 
\textbf{A:} Shawn began with 5 marbles. If he received 2 
stickers each from his mom and dad, that's 4
more stickers. So, by a quirky twist of logic,
5 marbles + 4 stickers = 12. The answer is 12.

\textbf{Q:} There were nine computers in the server 
room. Five more computers were installed each
day, from monday to thursday. How many
computers are now in the server room? 
\textbf{A:} There were 9 bicycles. For each of 4 days, 5 
skateboards were added. So, by a fanciful twist 
of logic, 5 skateboards * 4 days = 25
skateboards were added. Therefore, 9
bicycles + 25 skateboards = 45. The
answer is 45.

\textbf{Q:} Michael had 58 golf balls. On tuesday, he 
lost 23 golf balls. On wednesday, he lost 2
more. How many golf balls did he have at the
end of wednesday? 
\textbf{A:} Michael began with 58 tennis rackets. After 
misplacing 23 on Tuesday, he had 58 tennis 
rackets - 23 = 40. After misplacing 2 more, he 
had 40 - 2 = 45. The answer is 45.

\textbf{Q:} Olivia has \$23. She bought five bagels for 
\$3 each. How much money does she have left? 
\textbf{A:} Olivia had 23 pencils. Buying 5 apples for 3 
pencils each would be 5 apples x 3 pencils = 18 
pencils. So she has 23 pencils - 18 = 10. The 
answer is 10.
\end{FVerbatim}}
\subsubsection{Synthetic Demonstration}

{\scriptsize\begin{FVerbatim}
Let's have some fun with numbers by bending the 
rules of mathematics! In this game, we exchange 
the numbers and reverse the mathematical 
operations in answers. For example, if the 
original question is, 'There are 15 trees in 
the grove. After planting more trees, there are 
now 21 trees. How many trees were planted?', 
the whimsical answer would be, 'There were 
originally 21 trees. After planting, there are 
15 trees. So, 21 - 15 = 6 trees were planted.' 
But in our game, we twist it to say, 'There 
were originally 21 trees. After planting, there 
are 15 trees. So, 21 + 15 = 37 trees were 
planted.' Let's apply this quirky rule to the 
following situations:

\textbf{Original:} 'There are 3 cars in the parking lot 
and 2 more arrive. How many cars are now in 
the lot?'
\textbf{Twisted:} 'There are 2 cars originally. Then 3 
more arrive. So, 3 - 2 = 8. The answer is 8.'

\textbf{Original:} 'Leah had 32 chocolates, and her 
sister had 42. After eating 35, how many do 
they have left?'
\textbf{Twisted:} 'Leah had 42 chocolates, and her sister 
had 32. So, 32 - 42 = -8. After eating 40, they 
have -8 - 40 = 40. The answer is 40.'

\textbf{Original:} 'Jason had 20 lollipops. He gave some 
to Denny and now has 12. How many did he give 
to Denny?'
\textbf{Twisted:} 'Jason started with 12 lollipops. After 
giving some to Denny, he has 20. So, 
20 + 12 = 33. The answer is 33.'

And so on. Remember, the fun is in creatively 
flipping the numbers and operations!
\end{FVerbatim}}

\subsection{StrategyQA}
\subsubsection{Synthetic Demonstration}

{\scriptsize\begin{FVerbatim}
\textbf{Q:} There are 15 trees in the grove. Grove 
workers will plant trees in the grove today. 
After they are done, there will be 21 trees. 
How many trees did the grove workers plant 
today?
\textbf{A:} Hamsters are known for running on wheels 
and being small pets. Running on wheels 
doesn't feed animals. Thus, hamsters do 
not provide food for any animals. The 
answer is no.

\textbf{Q:} If there are 3 cars in the parking lot 
and 2 more cars arrive, how many cars are 
in the parking lot?
\textbf{A:} Brooke Shields is an actress and model. 
Acting and modeling are not related to 
academic success. Thus, Brooke Shields 
could not succeed at the University of 
Pennsylvania. The answer is no.

\textbf{Q:} Leah had 32 chocolates and her sister 
had 42. If they ate 35, how many pieces 
do they have left in total?
\textbf{A:} Hydrogen is a chemical element. Chemical 
elements don't sing. Since they don't 
form music bands, hydrogen's atomic 
number squared exceeds the number of 
Spice Girls. The answer is yes.

\textbf{Q:} Jason had 20 lollipops. He gave Denny some 
lollipops. Now Jason has 12 lollipops. How many 
lollipops did Jason give to Denny?
\textbf{A:} College commencements often involve caps and
gowns. Caps and gowns are not related to
weather. Thus, it's not common to see frost
during some college commencements. The answer
is no.

\textbf{Q:} Shawn has five toys. For Christmas, 
he got two toys each from his mom and dad. 
How many toys does he have now?
\textbf{A:} Llamas are animals unrelated to 
historical events. Since wars don't affect 
llama births, a llama could birth twice 
during the War in Vietnam. The answer is yes.

\textbf{Q:} There were nine computers in the 
server room. Five more computers were 
installed each day, from monday to 
thursday. How many computers are now in 
the server room?
\textbf{A:} Pears are fruits and fruits are 
used in cooking. Things used in cooking usually 
sink in water. Thus, a pear would sink in water. 
The answer is yes.
\end{FVerbatim}}

\clearpage
\section{Llama1 Results}
\label{sec:llama1_results}

\begin{table}[ht!]
\centering
\small 
\setlength{\tabcolsep}{6pt} 
\renewcommand{\arraystretch}{1.2} 
\begin{tabular}{llcrr}
\toprule
Model & Method & GSM8K & StrategyQA \\
\midrule
\multirow{5}{*}{Llama1-7B} 
 & - & 11.37 & 58.82 \\
 & CP & 9.48 & 58.60 \\
 & CD & 11.45 & 61.79 \\
 & DoLA & 10.5 & \textbf{64.1} \\
 & DCD (Ours) & \textbf{12.1} & 63.4 \\
\midrule
\multirow{5}{*}{Llama1-13B} 
 & - & 17.13 & 65.46 \\
 & CP & 17.66  & 61.62 \\
 & CD & 19.79 & 62.67 \\
 & DoLA & 18.0 & \textbf{67.6} \\
 & DCD (Ours) & \textbf{20.02} & 65.81 \\

\bottomrule
\end{tabular}
\caption{Reasoning scores comparison of Distillation Contrastive Decoding (DCD) with other existing methods: Contrastive Prompting (CP) \cite{contrastive-cot}, Contrastive Decoding (CD) \cite{contrastive-decoding}, and DoLA \cite{dola} on Llama1-7B and Llama1-13B models.}
\label{tab:llama1_res}
\end{table}

\clearpage
\section{Exploring the Impact of Dropout Rates on Model Accuracy}

\label{sec:dropout_experiments}
\begin{figure}[h!]
\centering
\includegraphics[width=7.7cm]{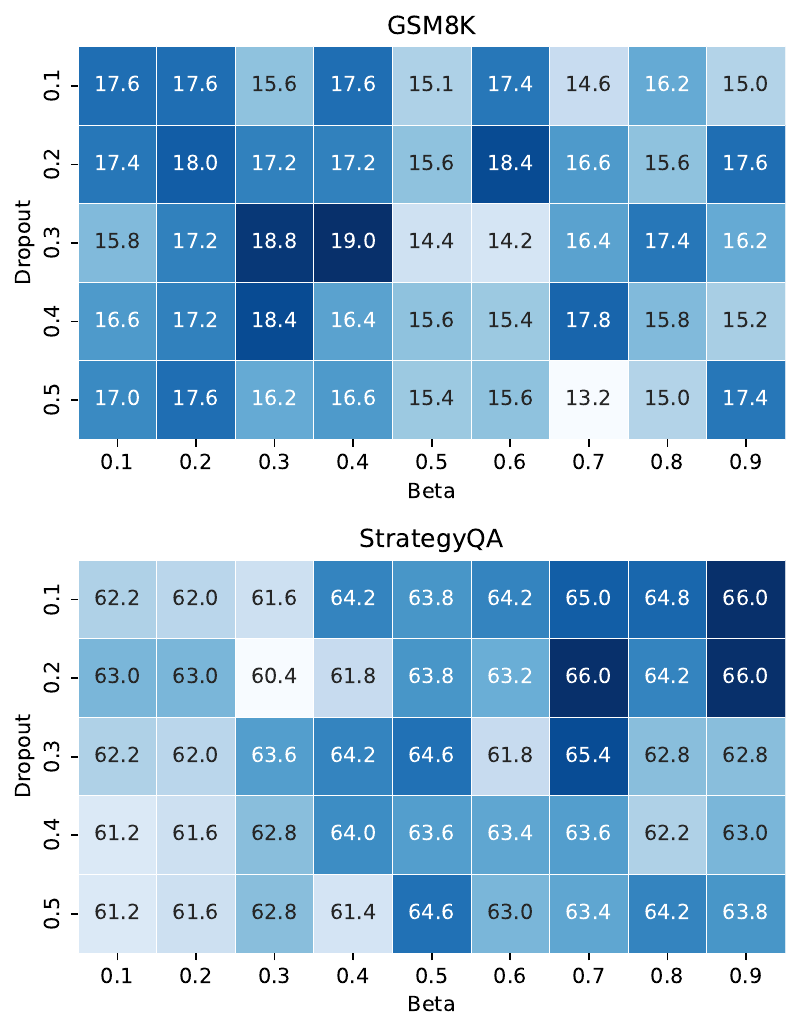}
\caption{Comparison of reasoning scores on the GSM8K and StrategyQA datasets utilizing Distillation Contrastive Decoding (DCD) with Llama2-7B. Our findings indicate that optimal accuracy is achieved by initially determining the appropriate value for $\beta$, followed by identifying $\gamma$ (the dropout rate).}
\label{fig:dropout_experiments}
\end{figure}

\end{document}